\documentclass[draft]{book}
\usepackage[FINAL]{fg}    

\title{FG 2006:\\
The 11th conference on Formal Grammar\\
Malaga, Spain\\July 29-30, 2006}
\shorttitle{FG-2006}
\author{Organizing Committee:\and
  Paola Monachesi\and Gerald Penn\and Giorgio Satta\and 
Shuly Wintner}
\date{}

\usepackage[comma]{natbib}  

\usepackage{chapterbib}     

\usepackage{import}         

\usepackage{makeidx}        

\usepackage{cslipubs-extra}

\usepackage{lingmacros}

\newtheorem{deriv}{Derivation}
\input{qobitree}
\usepackage{amssymb}
\usepackage{amsfonts}
\usepackage{amsmath}
\usepackage{pdfsync}
\usepackage{underbracket}

\begin{document}
\frontmatter
\maketitle        

\mainmatter       

\achapter[Treating clitics with MG]{Treating clitics with Minimalist Grammars}{Maxime Amblard}

\begin{abstract}
We propose an extension of Stabler's version of clitics treatment for a wider coverage of the french language. For this, we present the lexical entries needed in the lexicon. Then, we show the recognition of complex syntactic phenomenae as (left and right) dislocation, clitic climbing over modal and extraction from determiner phrase. The aim of this presentation is the syntax-semantic interface for clitics analyses in which we will stress on clitic climbing over verb and raising verb.
\end{abstract}


Minimalist Grammars (MG) is a formalism which was introduced in \cite{St97}, based on the Minimalist Program, \cite{Chom95}.
The main idea which is kept from the Minimalist Program is the introduction of constituent move in the formal calculus. Such a ``move" operation introduces flexibility in a system which seems to be like Categorial Grammars (CG). We try to recover the correspondance in CG, between syntactic structures and logical forms (interpretative level of the sentence).

This formalisation introduces constraints on the use of move rules, and by this way makes the syntactic calculus decidable. These grammars are lexicalised and all steps of the analysis are triggered by the information extracted from the lexicon: from a sentence, it selects a subset of words. To each word corresponds a sequence of features, and it is the first element of the sequence in the derivation which triggers the next rules.

An advantage of this system is that the structure of the calculus is constant. The coverage of the grammar is extended by adding new elements to the lexicon, never by adding new structural rules. The structural system of these grammars contains only two kinds of rules: move and merge (but extensions exist for both).
We refer the reader to Stabler's articles and others for presentation of the use of MG, \cite{St97}, \cite{VW99}.\\
\indent

Clitics are the normal form for pronoun in romance language. The syntactic and semantic behavior of clitics in these languages are complex. For French, clitics often climb over auxiliary verb. 
Ed Stabler proposes in \cite{St01} a partial lexicon for french clitics recognition and analysis.

We propose here to extend this lexicon to several well-known linguistic problems.
These problems interfere at different levels of analysis. Subject raising is typically a semantic question whereas the clitic climbing over modals is a syntactic question.
We propose a new lexicon for its syntactic analysis and then we will show how our semantic interface solves semantic questions.

We use the description of clitics proposed by Perlmutter in \cite{Pm71}. He proposes a filter to recognize the right order of clitics for romance languages, from where we extract the subfilter:
\begin{center} 
$[\{ je/tu/\cdots\} | ne | \{ me/te/se/\cdots\} | \{le/la/les/\cdots\} | \{lui/leur\} | y | en]$.\\
$[$nominative $|$ negative $|$ reflexive $|$ accusative $|$ dative $|$ locative $|$ genitive$]$.
\end{center}

\vspace{2ex}
\indent In the first part, we propose an extension of Stabler's version of clitics treatment for a wider coverage of the french language.  For this, we will present the lexical entries needed in the lexicon. Then, we will show the recognition of complex syntactic phenomena as (left and right) dislocation, clitic climbing and extraction from determiner phrase. The aim of this presentation is the last part: the syntax-semantic interface for clitics analyses in which we will stress on clitic climbing over verb and raising verb.

\section{Lexicon for french clitics}
\subsection{Stabler analysis}

Stabler's works on clitics are inspired by Sportiche \cite{Sp92},
who proposes the following treatment:

Clitics are not elements moved from position XP$^*$, but are coreferent with this position.
The clitics appearing in the structure bear all the features their co-refering XP$^*$ would bear.
Furthermore, clitics do not form an autonomous syntactic object, but they are built into a unit with some host.

In this work, two parts in the cliticization are distinguished.
The first one is an empty element which takes an argumental position from the verb. The second is the phonological treatment of the unit - the clitic in the surface structure.

We introduce lexical entries which are phonologically empty but carry special features which need to be unified with features of the phonological part of the clitic. 
The two different parts are connected by a move operation.
If just one of these items occurs in the sentence, derivation fails.

We sum up this treatment in the derivation as follow - the annotation recall the main feature of the word and the annotation on the $\epsilon$ recall the word which $\epsilon is the trace$:

 \enumsentence{
donne $\epsilon_{-F}$

Jean$_{-k} $ la$_{+F}$ donne $\epsilon_{-F}$ $\Rightarrow$ Jean$_{-k} $ la donne $\epsilon_{la}$

t$_{\epsilon}$ Jean$_{-k} $ la donne $\epsilon_{la}$

Jean t$_{\epsilon}$ $t_{Jean}$ la donne $\epsilon_{la}$.

John t$_{\epsilon}$ $t_{Jean}$ it gives $\epsilon_{it}$.

John give it.
}

In more details, the derivation is the following:

\begin{deriv}
Derivation of the simple french sentence : 
\emph{Jean la  donne}.

Lexicon:
\begin{center}
$\begin{tabular}{|ll|ll|ll|llll}
\hline
Jean & D -k  &$\epsilon$ & =T C & $\epsilon$ & D -k -G\\
 donne & V  & $\epsilon$&=$>$V =D +k =D v&&\\
 $\epsilon$ & =Acc3 +k T  & la & =v +G Acc3&&\\
\hline
\end{tabular}$
\end{center}
Derivation step by step:
\begin{enumerate}
\item selection of lexical entry : [ donne :: V]
\item selection of lexical entry : [$\epsilon$ :: =$>$V =D +k =D v] (which adds the syntactic component to the verb).
\item head movement. This is a merge between the two previous element where the phonological part of the argument moves to the phonological part of the head.
\item selection of lexical entry : [$\epsilon$ :: D -k -G]. This is the empty argumental verb position.
\item merge.
\item There is a licensee ``k" in first position, a move operation is triggered. After this step, the derivation tree is :

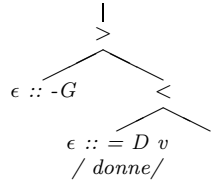
\begin{center}
\begin{footnotesize}
\leaf{$\epsilon$ :: -G }
\leaf{$\epsilon$ ::  = D v \\ / donne/}
\leaf{ $\quad$}
\branch{2}{$<$}
\faketreewidth{\hspace*{15mm}}
\branch{2}{$>$}
\branch{1}{}
\tree
\end{footnotesize}
\end{center}

\item selection of lexical entry : [Jean :: D -k].
\item merge.
\item selection of lexical entry : [la :: =v +G Acc3], the clitic takes part in the derivation.
\item merge.
\item move : the feature in the empty argument of the verb and the feature in the clitic  are cancelled.
\item selection of lexical entry : [$\epsilon$ :: =Acc3 +K T] - to the end of the derivation.
\item merge.
\item move : resolution of nominative case :
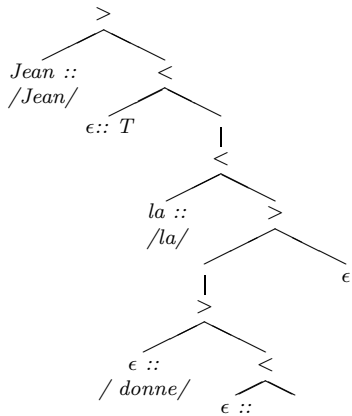
\begin{center}
\begin{footnotesize}

\leaf{Jean :: \\ /Jean/}
\leaf{ $\epsilon$:: T}
\leaf{ la :: \\ /la/}
\leaf{$\epsilon$ :: \\ / donne/}
\leaf{$\epsilon$ :: }
\leaf{$\quad$}
\branch{2}{$<$}
\branch{2}{$>$}
\branch{1}{}
\leaf{$\epsilon$}
\faketreewidth{\hspace*{2mm}}
\branch{2}{$>$}
\faketreewidth{\hspace*{15mm}}
\branch{2}{$<$}
\branch{1}{}
\faketreewidth{\hspace*{15mm}}
\branch{2}{$<$}
\faketreewidth{\hspace*{15mm}}
\branch{2}{$>$}
\tree\\[2ex]
\end{footnotesize}
\end{center}
\item selection of lexical entry : [$\epsilon$:: =T C] - empty ``complement" position.
\item merge ; end of the derivation with feature 'c' : acceptance.
\end{enumerate}
\end{deriv}

In his presentation, Stabler proposes a lexicon for accusative, dative and reflexive clitics recognition.
He ensures the right order with several verbal types. The analysis is driven by the head and the next clitics to introduce will have to be assigned verbal type as they occur in the Perlmutter filter's order.
Stabler uses the SMC - shortest move condition - to exclude the use of a reflexive and an accusative clitics in the same sentence.

\subsection{Extension: genitive, oblique and nominative clitics}
We can extend this first approach of french clitics treatment to other cases, in particular genitive, oblique and nominative. This section will present the lexical entries and the process of acceptance of derivations.

We call ``state of a verb" the basic type of the head currently handled. For example, if a verb has a accusative clitic its type will be ``Acc".

For genitive and oblique clitics, we just add in the lexicon two new empty argumental positions and a list of possible types for each clitic.

In a first time, we introduce a new verbal type for beginning the cliticization and another where the cliticization is finished.
We call them ``clitic" and ``endclitic".

Following the Perlmutter filter \cite{Pm71}, the first clitic we have to treat for keeping the right order is the genitive one.
We add a genitive state which is connected to the ``clitic" state.
The verbal state passes to  the genitive state by means of a lexical entry the phonological form of which is ``en" and carries a licensee feature``en":
$$[en]::[clitic<=, +EN, genitif].$$

From this state we pass to all the other states of the cliticization, for example :
$$[le]::[genitif<=, +G, acc].$$

and if there is only a genitive clitic, we use phonologically empty entry to pass to the end of the cliticization.
$$[]::[genitif<=, endclitic].$$

The ``oblique" clitics are treated in the same way, except that from ``oblique" it is impossible to go back to ``genitive".
All lexical entries of this type have a ``y" phonological form.
$$[y]::[clitic<=, +Y, oblique].$$
$$[y]::[genitive<=, +Y, oblique].$$

In the same way, from oblique we can pass to other possible clitic states, as for example :
$$[le]::[oblique<=, +G, acc].$$
$$[leur]::[oblique<=, +F, dat].$$
$$[]::[oblique<=, endclitic].$$

The nominative case is treated the same way. But the use of this procedure to add new clitic treatment is quadratic in the number of lexical entries.
For the nominative pronoun, a discussion could be opened around its clitic state. We consider here that they are clitics.

Another discussion about negative form could rise around the status of the negation marker whose position is after the pronoun.

For the moment, we do not treat the negative form in a right way so we will not include it in this presentation, but we assume that the treatment of nominative clitics is outside the clitic cluster. All the phonological pronoun entries take a verbal form in ``endclitic" state and give a  new verbal form in ``Nom"(inative) state.

We add an empty verb argument which must be included in the derivation before the clitic treatment:
$$[]::[d, -Subj, -case].$$

The sketch of the analysis is:

\begin{itemize}

\item la  donne $\epsilon_{-Nom}$ $\epsilon$

\item Je$_{+Subj}$ la  donne $\epsilon_{-Nom}$ $\epsilon$

\item Je $t_{Je}$ la donne $\epsilon$\\
I $t_{I}$ it give $\epsilon$\\
 I give it
\end{itemize}

We add in the lexicon a basic feature ``Nom" and the lexical entries of the nominative pronouns, for example:
$$[je]::[=endclitic, + Subj, Nom].$$
$$[nous]::[=endclitic, + Subj, Nom].$$

The derivation continues with a phonologically empty entry at the end of the derivation.
$$[]::[=nom, +case, t].$$

\section{Recognition of complex phenomena}
This treatment of french clitics is simple and can be integrated easily into a larger analysis.

\subsubsection{climbing over modal}
We treat the clitic climbing over the whole verbal cluster in particular over modal.

The modal is combine with the verb in the inflexion step.
The inflexion is treated with head movement and all clitics take their own place after this treatment.

If there are words which must be inserted between the verb and the modal - for sentences with adverbs - we first build the verbal constituent after which we treat the clitics. In this situation, the clitics could climb over the verb constituent or stand after.

For example, in french we can analyse a sentence as:
\enumsentence{
 Je l'ai vu.
  
  I him have seen.
  
  I have see him.
        }
by building the constituent \underbracket{ai vu}.
We can extend to sentences with inserted word: ``Je l'ai souvent vu" / ``I have often see him" 
with a derivation as :

\enumsentence{
 \underbracket{ai souvent vu} $\epsilon_{-Nom}$ $\epsilon_{-F}$

l'$_{+F}$ \underbracket{ai souvent vu} $\epsilon_{-Nom}$ $\epsilon_{-F}$ $\rightarrow$ l'\underbracket{ai souvent vu} $\epsilon_{-Nom}$ $\epsilon$

Je$_{+Nom}$ l'\underbracket{ai souvent vu} $\epsilon_{-Nom}$ $\epsilon_{-F}$ $\rightarrow$ Je l'\underbracket{ai souvent vu} $\epsilon$ $\epsilon$

I it $\quad$ often sawn

I often sawn him.
 }

\subsubsection{dislocation}
Clitic can be a direct recovery of a not-``empty verbal argument", for example in case of nominal dislocation.

There is a non empty verbal argument which must be extracted from the main sentence and become an indirect argument of the verb.

We build a verb with an ``argument which must be extracted" - a determiner phrase - DP - must be outside the main sentence. This state is introduced by a pause or comma. It modifies the determiner phrase in two different ways which depend on the side of the extraction:

\begin{itemize}
\item it adds a licensee for the left dislocation and cliticization.
\item it adds a licensee for cliticization (and nothing for right dislocation).
\end{itemize}

The main problem is to include in the sentence the right part which will be replaced by the clitic.

Left dislocation: the DP is extracted from the sentence, placed in first position and recovered by a clitic.
 \enumsentence{
  Marie le$_i$ voit trop \underbracket{ce type$_i$,} $\rightarrow$ Ce type$_i$, Marie le$_i$ voit trop.
 
That guy, Marie him sees too much.
 }

Lexical entry of modifier of DP.
$$[,]::[=>d,d, -H, -disloc].$$

Remark that we use coma to caring this treatment, but it can be through an empty lexical entry. The analyse would be the same.
The comma will be placed after the DP by a head movement.
The first licensee will be cancelled with the licensor of the clitic and the second with another entry that we must add in the classical ``comp" entry (this last entry is used to finish the derivation).
 
$$[]::[=t, c, +DISLOC].$$

Right dislocation :
In this case the determiner phrase is placed at the end of the sentence.
For the homogeneity of the mechanism, we add a licencee of recovered by a clitic, and another for the extraction at the end of the sentence.
$$[,]::[d<=,d, -H, -disloc].$$
	
The ``comp" phase uses a weak move which lets the phonological form of the constituent in its place - here, at the end of the sentence.

 \enumsentence{
        Marie le$_i$ voit trop \underbracket{, ce type$_i$}. $\rightarrow$ Marie le$_i$ voit trop, ce type$_i$.
        
        Marie him sees too much, that guy.
        }

This extraction seems to be very similar to questions:
in questions, an argument of the verb is extracted to take another position in the surface level of the sentence.

\subsubsection{Extraction from DP}
With the same kind of mechanism, we can extract an argument of any constituent. 
The determiner phrase can be complex and we extract an argument of the DP.
For example:

 \enumsentence{
Pierre en voit la fin - (Pierre voit la fin du film).

Peter of-it sees the end - Peter sees the end of the movie. 
}

We build ``la fin $\epsilon_{-en}$" and the cliticization allowed the extraction of the genitive.
``Pierre en voit la fin."

\subsubsection{Raising verb}

Raising verbs are verbs where one of the arguments is a verb and one of the other arguments is shared by both verbs, like in the sentence:
\enumsentence{Il semble le lui donner.
  
  He seems it him give.
  
  He seems give it to him.}

where the pronoun ``Il" is subject of the two verbs ``semble" and ``donner". The second verb must be in infinitive form.

In this case, the sentence has the following structures: 

[ subject raising\_verb clitic infinitive\_verb ].

A raising verb takes as an argument a verb in infinitive form - with a special inflexion ``infinitive" - and without subject.
The infinitive inflexion has the lexical entry: 

\begin{center}[-inf]::[=$>$v, verbe].\end{center}
``verbe" is the feature needed before starting the clitic treatment. 
A verbal form gets a ``verbe" type after the verb receives its inflexion.

The raising verb selects such a ``verb", then a DP subject and then becomes a VP of type ``raisingv" which means a VP which has not yet received the inflexion feature and will be able to receive new clitics (in particular pronoun).

For example:
\begin{center}
[semble]::[=verbe, =d, raisingv].
\end{center}

This verb should receive its inflexion and its subject. It follows this mechanism until the end of the derivation:

\vspace{-3ex}
\begin{verse}
\noindent
\begin{itemize}
\item semble \underbracket{la r\'epare-inf}
\item semble -$\epsilon$ \underbracket{la r\'epare-inf}
\item Je semble -$\epsilon$ \underbracket{la r\'epare-inf}

$\quad\quad\quad$ I seem -$\epsilon$ \underbracket{it repare-inf}

$\quad\quad\quad$ I seem repare it
\end{itemize}
\end{verse}

\section{Semantic interface}
\subsection{How to use the syntax/semantic interface}

From a sentence, we build a formula of higher order logic which represents its propositional structure.
We associate to each lexical entry a $\lambda$-term and to each syntactic rule an equivalent semantic rule.
We assume that the syntactic analysis drives the semantic calculus.

$\lambda$-terms application occurs only when an element has no features.
We assume the following functions:

$$\textit{feat}(x)=\left\{\begin{array}{l}1\;\textit{if\;the\;number\,\,of\,\,feature\,\,of}\;x=0\\0\;\textit{else}\end{array}\right.$$

$$\textit{sem}(x,y)=\left\{\begin{array}{l}1\; \textit{if \;feat(x) = 1 \;or\; feat(y) = 1}\\0\;\textit{else}\end{array}\right.$$

 Syntactic and semantic synchronisation:
after any operation in the syntactic calculus, the semantic counter part computes the \emph{sem} function and if $sem(x,y) = 1$, we perform the functionnal application of the two $\lambda$-terms. 
To known which application to perform, we look at the type of the semantic terms.

A semantic tree represents the semantic counter part of the sentence. It is a tree where the leaves are the semantic part of the lexical entries and the inner nodes contain the $\lambda$-term built and the direction of the head (of the syntactic part).
We use the following notation:
\begin{itemize}
\item breaker between direction head and $\lambda$-term : $\vdash$.
\item application: @
\end{itemize}

Applications are carried out when syntax allows it, therefore when the function $sem = 1$ for one of the two terms.
The following applications are possible:

\vspace{-2ex}
$$\begin{array}{lll}
$if sem ($\lambda$-term 1, $\lambda$-term 2) = 1$ & &$else$ \\

\leaf{$\lambda$-term 1}
\leaf{$\lambda$-term 2}
\branch{2}{$> \vdash \lambda$-term 1@ $\lambda$-term 2}
\tree
&
&
\leaf{$\lambda$-term 1}
\leaf{$\lambda$-term 2}
\branch{2}{$> \vdash \lambda$-term 1, $\lambda$-term 2}
\tree
\end{array}$$

If a move operation cancelled the last feature, we represent it by a unary branch in the tree.

Remark.
There are two different possibilities for the semantic calculus: 
either waiting for elements completely discharged either immediately perform the application. But both fail in different cases: immediate application fails in case of ``late adjunction" and the other possibility fails in questions treatment. The right solution seems to be intermediate: it consists in determining a subset of features which must be consumed before applications will be performed. For the moment, we choose the first possibility. Later on, we shall do differently but this only involve changes in the \emph{feat} function.

\subsection{Example of semantic treatment}
\subsubsection{Clitic semantics}

We present a syntactic treatment of clitics in two different parts. One is phonologically empty and is the non empty argument of the verb, the other is syntactically empty but it is a phonological recovery of the first one.
The semantical part of the clitic is in the argumental position and this is a free variable which must be bound in the context. The phonological recovery is an identity.

\begin{center}
$\begin{array}{|l|l|l|}
\hline
$lexical entries$ & $syntactic form$ & $semantic form$\\
la & dat<= \;+G \;acc  & Id \\
t(la) & p \;-case\; -G  & x^*\\
\hline
\end{array}$
\end{center}

* Free variable, bound in the context - we could use the Bonato algorithm to determinated how this variables are bounded, \cite{B06}

We briefly present a semantic tree for a clitic treatment: 

 \enumsentence{
 Jean la r\'epare.
 
John it repairs.

John repairs it.
}

In the semantic tree of the part of the clitisization above,  we do not represent the identity operator (except for the clitic one).

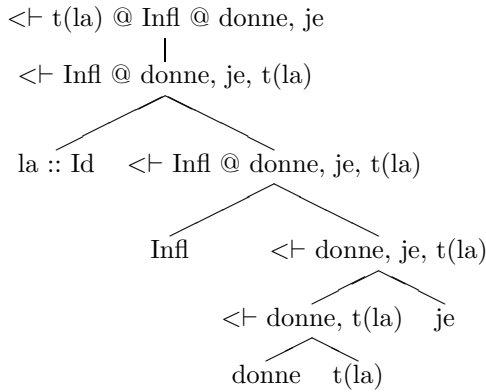
\begin{center}
\leaf{la :: Id}
\leaf{Infl}
\leaf{donne}
\leaf{t(la)}
\branch{2}{$< \vdash$ donne, t(la)}
\leaf{je}
\branch{2}{$< \vdash$ donne, je, t(la)}
\branch{2}{$< \vdash$ Infl @ donne, je, t(la)}
\branch{2}{$< \vdash$ Infl @ donne, je, t(la)}
\branch{1}{$< \vdash$ t(la) @ Infl @ donne, je}
\tree
\end{center}

The last part of the tree is built by a move which creates a link between the phonological part of the clitic and the argumental part.

\subsubsection{Over raising verbs}

For the semantic calculus, raising verbs are predicates which take a subject and an action as argument. They apply a variable at this action.

We present the analysis of the sentence: 
\enumsentence{
Je semble la r\'eparer.

I seem it repair.

I seem repair it.
}

The $\lambda$-terms, semantic counter-part of lexical entries are:

\begin{center}
\begin{tabular}{|l|l|}
\hline
sembler & $\lambda S \lambda v . (seem \;v , \; S(v))$\\
Je & I\\
$\epsilon_{la}$ & Y$^{*}$ \\
r\'eparer & $\lambda$ x $\lambda$ y . repair (y, x)\\
\hline
\end{tabular}
\end{center}
$\;^*$ this variable is bound in the context

The semantic counter part of the pronoun is a constant refering to the speaker ``I".
The clitic subject climbs over the raising verb. It can be the subject of both verbs in the sentence due to the semantic structure of the raising verb. If the main verb of the sentence has a subject, the application will not introduce a new variable in the formula, else the main verb needs a variable which stands at the subject place.  The raising verb involves this variable by duplication of its subject. 

The syntactic analysis builds the following structure:
$$ (I @ (inflexion @ (seem @ (la (infinitive @ repare))))) $$

which allows the computation of the formula:
``la reparer" 
$$ \lambda x . repair (x, Y) $$

and this term is applied to the raising verb: $\lambda S \lambda v . (seem \;v , \; S(v))$
$$ \lambda v . seem (v, repair (v, Y)) $$

At the end of the calculus, we construct the formula:
$$pres(seem (I , repair(I, Y)))$$ where Y is bound in the context.

This is the formula we want to construct for representing the propositionnal semantics of the sentence.
The subject clitic syntactically climbs over the main verb, and semantically climbs over the two verbs.

\section{Conclusion and future work}

In this paper, we presented an extension of Ed Stabler's propositions on french clitics in minimalist grammars.
The new lexicon makes it possible to treat several other syntactic phenomena, the same way as clitic climbing, e.g. extraction from NP or right and left dislocation.

Then, we proposed a syntax-semantic interface for Minimalist Grammars. The aim of this calculus is to build a formula of higher order logic. The semantic calculus - $\lambda$-calculus - is driven by the syntactic one. We emphasize on the way to recognize clitics and semantic implication of climbing with raising verbs.

For future work, we want to integrate the negation into the grammar. We consider that the neg-marker ``ne" is a clitic and must be incorporated in the treatment of french clitics.
There is another complex phenomenon to consider concerning with clitics in the imperative mode (and negation).

Other cases of raising verbs exist which are more complex, allowing several syntactic clitic climbings as in:
\enumsentence{
Je la laisse le lui donner.

I her let it (to) him give.

I let her  give it to him.
}

where clitics take place in different orders.

Moreover, we want to continue to modelize the semantic effect of clitics in sentences, in particular for interaction between quantifier scope and clitics, which can introduce ambiguities in sentences like:
\enumsentence{Je la laisse tous les lui donner.

I her let all them him give.

I let her gives all to him.}

\section*{Acknowledgement}

The writer like to thank Christian Retor\'e and Alain Lecomte for crucial supports and one of the anonymous TALN 2006 referees for important comments and examples reuse in this paper.


\bibliographystyle{cslipubs-natbib}
\nocite{*}
\bibliography{FG06ref}

\end{document}